\documentclass[twocolumn,letterpaper,10pt]{article}
\usepackage{float}
\usepackage{kuz}
\usepackage[round,authoryear]{natbib}
\usepackage{amsmath}
\usepackage[varg]{txfonts}   
\usepackage{hyperref}
\usepackage{graphicx}
\usepackage[utf8]{inputenc}
\usepackage{todonotes}

\usepackage[labelfont=bf]{caption}
\newcommand{\argmax}{\operatornamewithlimits{argmax}}

\graphicspath{{graphics/}{graphics/arch/}{Graphics/}{./}} 
\begin{document}
\title{Where is my forearm? Clustering of body parts from simultaneous tactile and linguistic input using sequential mapping
}
%
%
\author{
\vspace{1ex}
\textbf{Karla Štěpánová$^{1,2}$, Matěj Hoffmann$^{1,3}$, Zdeněk Straka$^1$, Frederico B. Klein$^4$, Angelo Cangelosi$^4$,  Michal Vavrečka$^{2}$}\\
$^{(1)}$ Center for Machine Perception, Department of Cybernetics, \\ Faculty of Electrical Engineering, Czech Technical University in Prague\\
$^{(2)}$ Czech Institute of Informatics, Robotics, and Cybernetics, CTU in Prague\\
$^{(3)}$ iCub Facility, Istituto Italiano di Tecnologia\\
$^{(4)}$ School of Computing, Electronics and Mathematics, Plymouth University, Plymouth, UK\\
}

\maketitle

\abstract{%
  Humans and animals are constantly exposed to a continuous stream of sensory information from different modalities. At the same time, they form more compressed representations like concepts or symbols. In species that use language, this process is further structured by this interaction, where a mapping between the sensorimotor concepts and linguistic elements needs to be established. There is evidence that children might be learning language by simply disambiguating potential meanings based on multiple exposures to utterances in different contexts (cross-situational learning). In existing models, the mapping between modalities is usually found in a single step by directly using frequencies of referent and meaning co-occurrences. In this paper, we present an extension of this one-step mapping and introduce a newly proposed sequential mapping algorithm together with a publicly available Matlab implementation. For demonstration, we have chosen a less typical scenario: instead of learning to associate objects with their names, we focus on body representations. A humanoid robot is receiving tactile stimulations on its body, while at the same time listening to utterances of the body part names (e.g., hand, forearm and torso).  With the goal at arriving at the correct ``body categories'', we demonstrate how a sequential mapping algorithm outperforms one-step mapping. In addition, the effect of data set size and noise in the linguistic input are studied.

}
\maketitle
\section{Introduction}
\label{intro}

Body representation has been the topic of psychological, neuroanatomical and neurophysiological studies for many decades. Spurred by the account of \citet{Head1911} and their proposal of superficial and postural schema, a number of different concepts has been proposed since: body schema, body image, and corporeal schema being only some of them. Body schema is usually thought of as more ``low-level'', sensorimotor representation of the body used for action. Body image is an umbrella term uniting higher level representations, for perception more than for action, and accessible to consciousness. \citet{Schwoebel2005} amassed evidence for distinguishing between three types of body representations: body schema, body structural description, and body semantics---constituting a kind of hierarchy. The body structural description is a topological map of locations derived primarily from visual input that defines body part boundaries and proximity relationships. Finally, body semantics is a lexical–semantic representation of the body including body part names, functions, and relations with artifacts (e.g., shoes are used on the feet, and feet can be used to kick a football).

While the details of every particular taxonomy or hierarchy can be discussed, clearly, there is a trend from continuous, modality-specific representations (like the tactile homunculus) to multimodal, more aggregated representations. This may be first instantiated by increasing receptive field size and combining sensory modalities, as it is apparent in somatosensory processing, e.g. areas relatively specialized on proprioception or touch and with small receptive fields (like Brodmann areas 3a and 3b), touch and proprioception are getting increasingly combined in areas 1 and 2. Then, going from anterior to posterior parietal cortex, the receptive fields grow further and somatosensory information is combined with visual.
One can then ask whether this process of bottom-up integration or aggregation may give rise to discrete entities, or categories, similar to individual body parts. \citet{Vignemont2009} focused on how body segmentation between hand and arm could appear based on a combined tactile and visual perception. They explored category boundary effect which appeared when two tactile stimuli were presented: these stimuli felt farther away when they were applied across the wrist than when they were applied within a single body part (palm or forearm). In conclusion, they suggest that the representation of the body is structured in categorical body parts delineated by joints, and that this categorical representation modulates tactile spatial perception.

Next to the essentially bottom-up clustering of multimodal body-related information, an additional ``categorization'' of body parts is imposed through language, such as when the infant hears her parents naming the body parts. 
Interestingly, recent research~\citep{Majid2010} showed that there are some cross-linguistic variabilities in naming body parts and this may in turn override or influence the ``bottom-up'' multimodal (non-linguistic) body part categorization. 

While the field is relatively rich in experimental observations, the mechanisms behind the development and operation of these representations are still not well understood. Here, computational and in particular robotic modeling ties in---see \citep{Hoffmann2010,Schillaci2016} for surveys on body schema in robots. 
\citet{Petit2016} developed an algorithm for the iCub humanoid robot to associate labels for body parts and later proto-actions with their embodied counterparts. These could then be recombined in a hierarchical fashion (e.g., ``close hand'' consists of folding individual fingers). 
\citet{Mimura2017} used Dirichlet process Gaussian mixture model with latent joint to provide a Bayesian body schema estimation based on tactile information. Their results suggest that kinematic structure could be estimated directly from tactile information provided by a moving fetus without any additional visual information---albeit with a lower accuracy. Our own work on the iCub humanoid robot has thus far focused on learning primary representations---tactile \citep{HoffmannStraka2017} and proprioceptive \citep{HoffmannBednarova2016}. In this work, we use the former (the ``tactile homunculus'') as input for further processing---interaction with linguistic input.

In this work, we strive to find segmentation of body parts based on a simultaneous tactile and linguistic information.
However, body part categorization and mapping to body part names is one instance of a more general problem of segmenting objects from the environment, learning compressed representations (loosely speaking: concepts, categories, symbols) to stand in for them and associating them with words to which the infant is often exposed simultaneously. \citet{Borghi2004}, for example, studied the interaction of object names with situated action on the same objects.

We made use of a newly proposed sequential mapping algorithm which extends an idea of one-step mapping~\citep{Smith2006} and compared its overall accuracy to one-step mapping as well as to accuracies of segmenting individual body parts. We further explore how the accuracy of the learned mapping is influenced by a level of noise in the linguistic domain and data set size. The sequential mapping strategy was shown to be very robust as it can find the mapping under circumstances of very noisy input and clearly outperformed the one-step mapping.

Complete source code used for generating results in this article is publicly available at  \url{https://github.com/stepakar/sequential-mapping}.

This article is structured as follows. The inputs and their  preprocessing and the mapping algorithms are described in Section \ref{sec-1}. This is followed by Results (Section \ref{sec:results}) and a Discussion and Conclusion. 

\section{Materials and Methods}
\label{sec-1}
In this section, we will first present the inputs and their preprocessing pipelines: tactile input (Section \ref{sec:tactileInput}) and linguistic input (Section \ref{sec:lingInput}). In total, 9 body parts of the right half of the robot's upper body were stimulated: torso/chest, upper arm, forearm, palm and 5 fingertips. Tactile stimulation coincided with an utterance of the body part's name. Then,  
the one-step and sequential mapping algorithms (sections \ref{sec:oneStepM} and \ref{sec:SM}) are presented, and a description of the evaluation (Section \ref{sec:evaluation}).

\subsection{Tactile inputs and processing}
\label{sec:tactileInput}
\begin{figure*}[!t]
\centering
\includegraphics[width=\textwidth,clip]{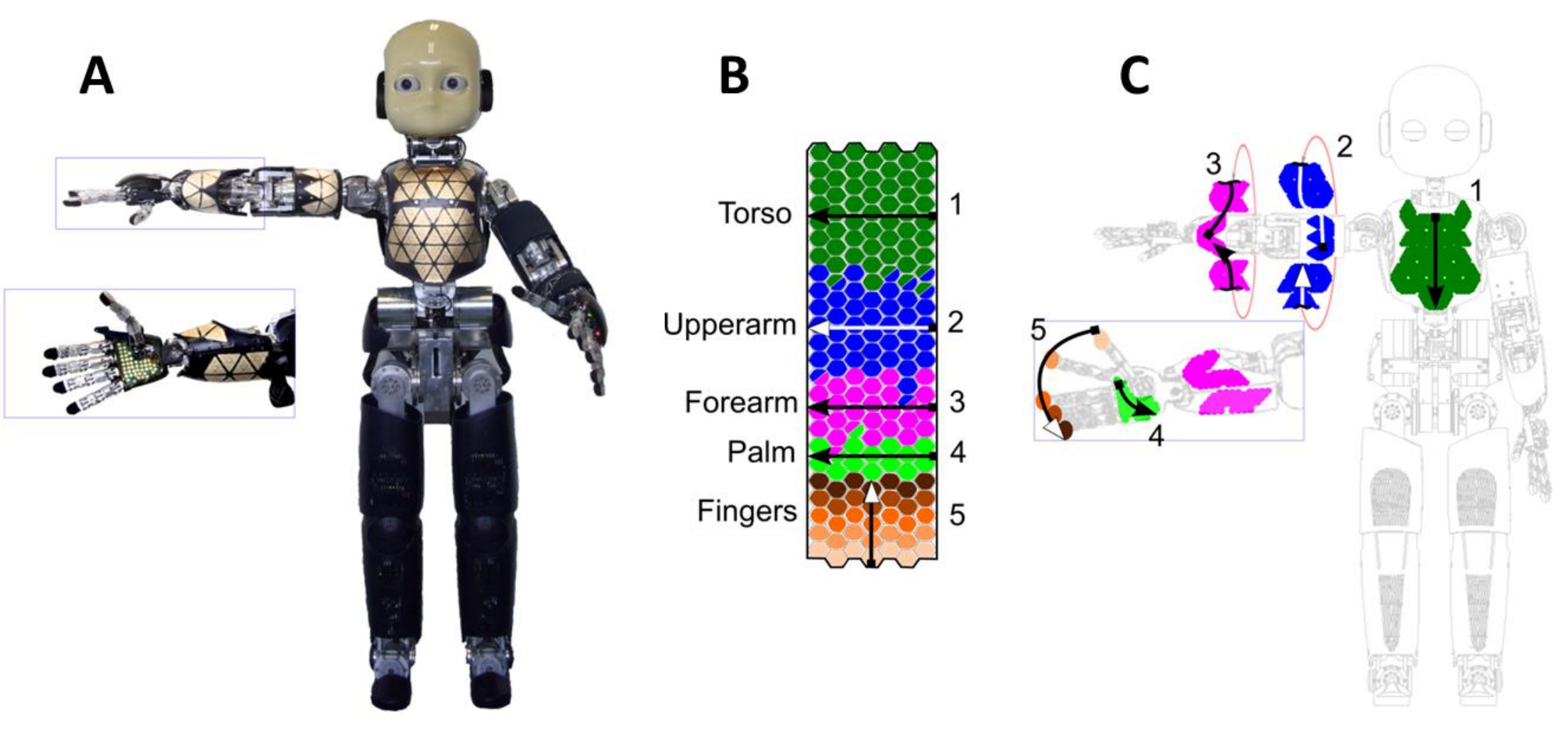}
\caption{ iCub skin and tactile homunculus. (A) Photograph of the iCub robot with artificial skin exposed on the right half of the upper body (1154 taxels in total). (B) Representation of tactile inputs learned using a Self-Organizing
Map – a $24 \times 7$ neuronal sheet. (C) Schematics with skin patches unfolded and
colored to mark the correspondence with B. Arrows illustrate the relationship
in orientation between skin parts and the learned map~\citep{HoffmannStraka2017}.}
\label{fig:icubHomun}     
\end{figure*}

To generate tactile stimulation pertaining to different body parts, we built on our previous work on the iCub humanoid robot. In particular, the ``tactile homunculus'' \citep{HoffmannStraka2017}---a primary representation of the artificial sensitive skin the robot is covered with (see Fig. \ref{fig:icubHomun} -- one half of the robot's upper body). In the current work, the skin was not physically stimulated anymore, but the activations were emulated and then relayed to the ``homunculus'', as detailed below. 

\subsubsection{Emulated tactile input}
\label{sec:emulTact}
We created a YARP \citep{Metta2006} software module to generate virtual skin contacts\footnote{\url{https://github.com/robotology/peripersonal-space/tree/master/modules/virtualContactGeneration}}. A skin part was randomly selected and then stimulated. The number of pressure-sensitive elements (henceforth taxels) for different skin parts was 440 for the torso, 380 for upper arm, 230 for forearm, and 104 for the hand (44 for palm and 5 $\times$ 12 for fingertips)---1154 taxels in total. Once the skin part was randomly selected, a small region was also randomly picked within that part for the tactile stimulation---10 taxels at a time, corresponding to the triangular modules the skin is composed of. For the hand, the situation was slightly different: the entire hand was treated as one skin part. Then, within the hand, a random choice was made between 5 subregions on the palm skin (8 to 10 taxels) and 5 fingertips (12 taxels each). Data was collected for 100 minutes, corresponding to approximately 2000 individual 3 second stimulations. For all skin parts, the stimulation lasted for 3 seconds and was sampled at 10 Hz. A label--body part name--was saved along with the tactile data. These labels are used to generate the linguistic input and for performance evaluation later, but do not directly take part in the clustering of tactile information. Please note that there were separate labels for the palm and individual fingers, while these were all treated as one ``skin part'' in the virtual touch generation and hence the number of samples per finger, for example, was lower than for other non-hand body parts.

\subsubsection{First layer -- ``tactile homunculus''}
The input layer of the ``tactile homunculus'' \citep{HoffmannStraka2017} consists of a vector, ${\bf a}(t)$, of activations of 1154 taxels at time $t$---the output of the previous section---that have binary values ($1$ when a taxel is stimulated, $0$ otherwise). The output layer then forms a $7 \times 24$ (168 ``neurons'' in total) grid -- see Figure~\ref{fig:icubHomun}~B. This layer is a compressed representation of the skin surface---the receptive fields of neurons (the parts of skin they respond to) are schematically color-coded. However, this code (and ``clustering'') is not available as part of the tactile input.

The output layer will be represented as a single vector ${\bf x}(t) = [ x_1(t), ..., x_{168}(t)]$.
The activations of the output neurons, $x_i(t)$, are calculated as dot products of the weight vector ${\bf u}_i$ corresponding to the $i$-th output neuron and the  tactile activation vector ${\bf a}(t)$ as follows: 
\begin{equation}
x_i(t)={\bf u}_i \cdot {\bf a}(t)
\end{equation}
\subsubsection{Second layer -- GMM}

The output of the first layer, vector ${\bf x}(t)$ (168 elements, continuous-valued)  serves as input to the second tactile processing layer. This layer aims to cluster individual body parts and represent them as abstract models. Resulting models $T_j$ are subsequently mapped in the multimodal layer to clusters found in the language layer.

To process the outputs from the first layer, we used a Gaussian mixture model (GMM), which is a convex mixture of $D$-dimensional Gaussian densities $l({\bf x}|{\bf \theta}_j)$. In this case, each tactile model $T_j$ is described by a set of parameters ${\bf \theta}_j$. The posterior probabilities $p({\bf \theta}_j|{\bf x})$ are computed as follows:

\begin{equation}
p({\bf \theta}_j|{\bf x})=\sum \limits_{j=1}^{J}{r_j^k l({{\bf x}}|{\bf \theta}_j)},
\end{equation}
\begin{equation}
l({\bf x}|{\bf \theta}_j)={1 \over {\sqrt{(2 \pi)^{D}}\sqrt{|{\bf S}_j|}}}\exp [ -{1 \over 2}({\bf x}-{\bf m}_j)^T ({\bf S}_j)^{-1}({\bf x}-{\bf m}_j)],
\end{equation}
where ${\bf x}$ is a set of $D$-dimensional continuous-valued data vectors, $r_j^k$ are the mixture weights, $J$ is the number of tactile models, parameters ${\bf \theta}_j$ are cluster centers ${\bf m}_j$ and covariance matrices ${\bf S}_j$.   

Mixture of Gaussians is trained by the EM algorithm \citep{dempster1977}. Number of tactile models $J$ is in this model preset based on the number of different linguistic labels. In future, we plan to use an adaptive extension of GMM algorithm such as gmGMM~\citep{Stepanova2016} to detect this number autonomously.

An output of this layer for each data point ${\bf x}(t)$  is the vector ${\bf y}(t)$ of $J$ output parameters describing the data point (the likelihood that the data point belongs to each individual cluster in a mixture). This corresponds to the fuzzy memberships (distributed representation).

\subsection{Linguistic inputs and processing}
\label{sec:lingInput}
Tactile stimulation of a body part was accompanied with the  corresponding utterance. In our case, where we have 9 separate body parts, these are 'torso', 'upper arm', 'forearm', 'palm', 'little finger', 'ring finger', 'middle finger', 'index finger' and 'thumb'. Linguistic and tactile inputs are processed simultaneously.

We conducted experiments with spoken language input---one-word utterances pronounced by a non-native English speaker. 
To process this data, we made use of CMU Sphinx (an open-source flexible Markov model-based speech recognizer system) \citep{lamere2003} and achieved 100\% accuracy of word recognition. The word-forms are extracted from the audio input and compared to prelearned language models by means of the log-scale scores $p({\bf w}_t^n|L_i)$ of the audio matching. Based on these data, posterior probability can be computed. 

However, in the current work, we employed a shortcut and used the labels (ground truth) directly. 
This allowed us to fully explore the effect of misclassification in linguistic subdomain to mapping accuracy. 
The noise to the language data was added subsequently and evenly to all classes (a given proportion of labels was randomly permuted). 

\subsection{Cross-situational learning}
\label{sec:crossSit}
One possible way how to establish mapping between sensorimotor concepts and linguistic elements is to use frequencies of referent and meaning co-occurrences, that is, the ones with the highest co-occurrence are mapped together~\citep{Smith2006,Xu2007}. This method is usually called cross-situational learning and supposes the availability of the ideal associative learner who can keep  track and store all co-occurrences in all trials, internally memorizing and representing the word–object co-occurrence matrix of input. This allows the learner to subsequently choose the most strongly associated referent~\citep{Yu2012}. 

\subsubsection{One-step mapping}
\label{sec:oneStepM}
The simplest one-step word-to-referent learning algorithm only accumulates word-referent pairs. This can be viewed as Hebbian learning: the connection between a word and an object is strengthened if the pair co-occurs in a trial. To extend this basic idea, we can enable also forgetting by introducing a parameter $\eta$, which can capture the memory decay~\citep{Yu2012}. Supposing that at each trial $t$ we observe an object $o_{t}^n$ and hear a corresponding word $w_t^n$ ($N_t$ possible associations), we can describe the update of the strength of the association between word model $L(i)$ and object---in our case tactile model $T(j)$---as follows:

\begin{equation}
\label{eq:cooc}
A(i,j)=\sum_{t=1}^R \eta(t)\sum_{n=1}^{N_t} \delta(w_{t}^n,i)\delta(o_{t}^n,j),
\end{equation}
where $R$ is the number of trials, $\delta$ is the Kronecker delta function (equal to 1 when both arguments are identical and 0 otherwise), $w_t^n$ and $o_t^n$ indicate the $n$th word–object association that the model attends to and attempts to learn in the trial $t$ and $\eta(t)$ is the parameter controlling the gain of the strength of association.

Now let's assume that the word $w(i)$ is modeled by the model $L_i$ in the language domain and object (referent) $o(j)$ is modeled by the model $T_{m(i)}$ in the tactile domain. Our goal is to find the corresponding model $T_{m(i)}$ from tactile subdomain for each model $L_{i}$ from language domain  to assign them together. Indices $ m(i)$ are found as follows: 
	\begin{equation}
	\label{eq:OM}
	\forall {i} : m(i) = \argmax_i A(i,j)  ,
	\end{equation}
	where ${\bf A}$ is the co-occurrence matrix computed in the Eq.~\ref{eq:cooc} (element $A(i,j)$ captures co-occurrence between the word $w(i)$ and object $o(j)$). 
    
\subsection{Sequential mapping algorithm}
\label{sec:SM}

To capture dynamic competition among models, we extend the basic one-step mapping algorithm for cross-situational learning by sequential addition of inhibitory connections. The inhibitory mechanisms and situation-time dynamics were already partially included into the model of cross-situational learning proposed by \citet{McMurray2012}. Even though our model shares some similarities with the model proposed by McMurray, it stems from different computational mechanisms. After a reliable assignment between a language and tactile model is found, inhibitory connections among this tactile model and all other language models are added. Thanks to this mechanism, mutual exclusivity principle (the fact that children prefer mapping where object has only one label to multiple labels~\citep{Markman1989}) is guaranteed.

The assignment between tactile models $T_j$ and language models $L_j$ is found using the following iterative procedure:

\begin{enumerate}
\item Tactile and language data are clustered separately and the corresponding posterior probabilities are found.
\item For each data point the most probable tactile and language clusters are selected and the data point is assigned to these clusters.
\item Co-occurrence matrix with elements $A(i,j)$ is computed and the best assignment is selected:
\begin{equation}
\label{eq:sm}
[im,m(im)]=\argmax_i \argmax_j A(i,j).
\end{equation}
In this step, the tactile model $T_{m(im)}$ is assigned to the language model $L_{im}$. 
\item Inhibitory connections are added between the assigned tactile model $T_{m(im)}$ and all language models $L_i$, where $i \neq im$ (mutual exclusivity).
\item Assigned data points (data points which belong to both $T_{m(im)}$ and $L_{im}$) are deleted from the data set.
\item If data set is not empty or not all tactile clusters are assigned to some language cluster go to (1), else stop.
\end{enumerate}

\subsection{Evaluation}
\label{sec:evaluation}

Accuracy of the learned mapping is calculated in the following manner: We cluster output activations from the tactile homunculus and assign each data point to the most probable cluster. Then, we find indices~$m(i)$  for all clusters as defined in equation~\ref{eq:OM} for one-step mapping and equation~\ref{eq:sm} for sequential mapping. Based on this mapping we can assign each data point to the language label. These language labels are subsequently compared to the ground truth (the body part name is equivalent to the language label prior to the application  of noise). Accuracy is then computed as:
\begin{equation}
acc = TP/N
\end{equation}
where $TP$ (true positive) is the number of correctly assigned data points and $N$ is the number of all data points.

\section{Results}
\label{sec:results}
We studied the performance of one-step vs. the sequential mapping algorithms on the ability to cluster individual body parts from simultaneous tactile and linguistic input. That is, all the skin regions on the same body part should ``learn'' that they belong together (to the forearm, say), thanks to the co-occurrences with the body part labels. In addition, the effect of data set size and levels of noise in the linguistic domain are investigated (Section \ref{sec:OMvsSM}). A detailed analysis of the mapping accuracy for individual body parts and a backward projection onto the tactile homunculus are shown in sections \ref{sec:indBodyParts} and \ref{sec:backHomun} respectively.

\subsection{Comparison of accuracy of one-step mapping to sequential mapping}
\label{sec:OMvsSM}
\begin{figure*}[!ht]
\includegraphics[width=\textwidth,clip]{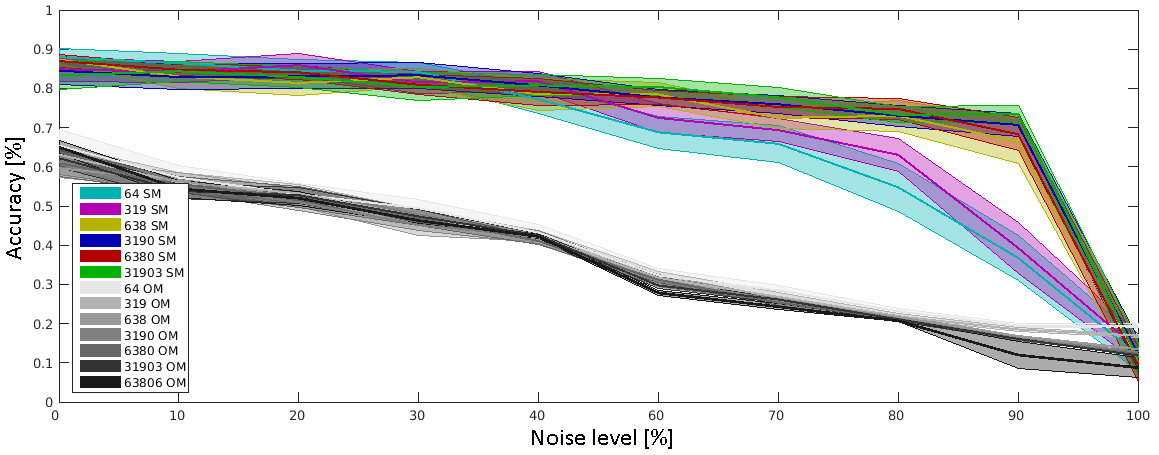}
\caption{Accuracy of one-step vs. sequential mapping for different levels of noise in language. Number denotes the size of data set, SM - sequential mapping and OM - one-step mapping. The mean and standard deviation from 20 repetitions is visualized.}
\label{fig:SMnoise}     
\end{figure*}
The performance of the one-step and sequential mapping algorithms is shown in Fig.~\ref{fig:SMnoise}. The comparison  is provided for different data set sizes (namely for 6 different data sets with number of data points from 64 to 63806) and noise levels. As can be seen, the accuracy of sequential mapping remains very stable and outperforms one-step mapping for all values of the noise (in the linguistic domain) and all data set sizes. For smaller data sets, we can see a steeper drop in accuracy with increasing noise in the language data.

\subsection{Accuracy of mapping for individual body parts}
\label{sec:indBodyParts}
\begin{figure*}[!ht]
\centering
\includegraphics[width=\textwidth,clip]{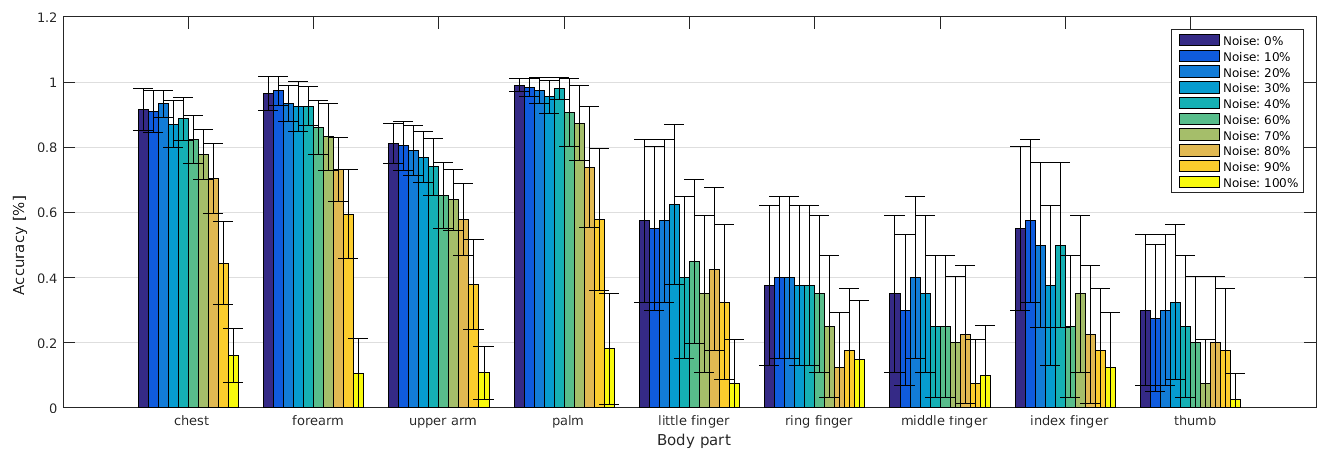}
\includegraphics[width=\textwidth,clip]{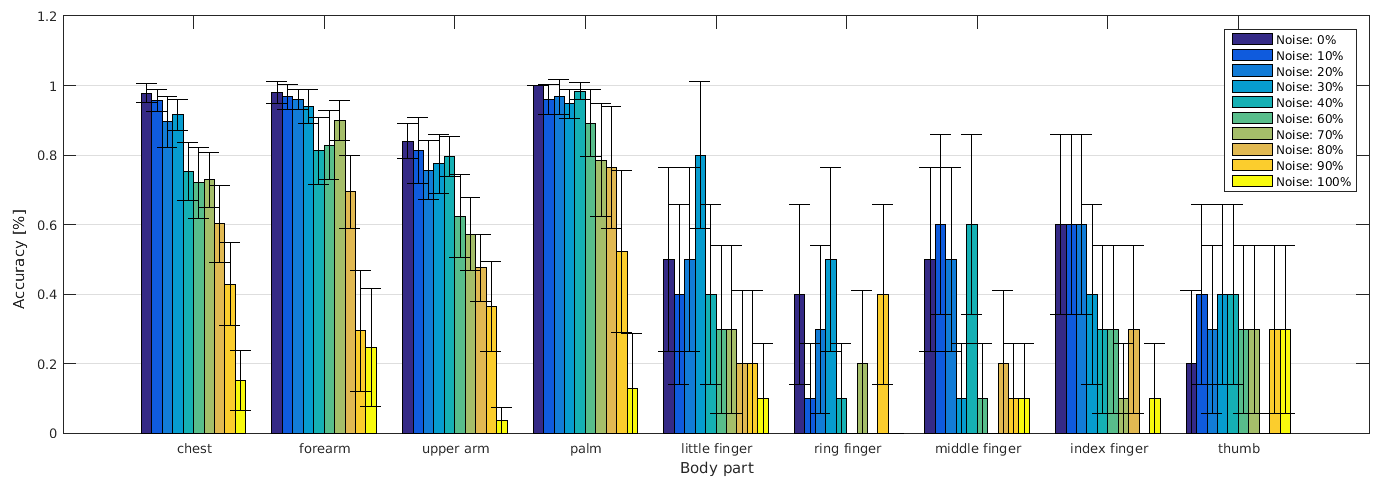}
\caption{Accuracy of sequential mapping for individual body parts: Visualization of sequential mapping accuracy based on the noise in linguistic data for 2 data set sizes: 3190 data points (upper) and 638 data points (lower) , noise in language data 0-100\%(random). The mean and standard deviation from 40 repetitions are visualized.}
\label{fig:BPnoise}     
\end{figure*}
The accuracy calculated in the previous section and Fig. \ref{fig:SMnoise} is an overall accuracy and we don't take into account the number of data points per individual body part. To explore the performance in more detail, we focused also on the accuracy of sequential mapping for individual body parts. The results for the data set with 3190 and 638 data points can be seen in Fig.~\ref{fig:BPnoise} top and bottom panel, respectively. The accuracy for all body parts decreases with increasing noise in the linguistic input. The accuracy for fingers is significantly lower---this is due to the lower number of samples per finger (see Section \ref{sec:emulTact}).
Comparing the top and bottom panel in Fig.~\ref{fig:BPnoise} demonstrates poorer performance with higher variance, especially for the fingers.  
\subsection{Projecting results of sequential mapping back onto homunculus}
\label{sec:backHomun}

After tactile data from homunculus are clustered and these clusters are mapped to appropriate language clusters (representing body parts utterances), we can project these labels back onto the original tactile homunculus. Considering that $x_i(t)$ are activations of neuron $i$ in the homunculus, $D$ is the whole data set consisting of vector of homunculus activations for each data point,  and $LangLabel(d)$ is the language label assigned to a data point $d$ based on the sequential mapping procedure described in the Section~\ref{sec:SM}, we can project results of sequential mapping onto the homunculus in a following manner. First, we compute strength of activation $n_i^k$ of each neuron $i$ for a given language label $k$ as follows:

\begin{align}
\centering
n_i^k = &\sum_{{\bf x}(t) \in D_k} x_i(t),\ i \in \{1,\dots,168\},\\ \nonumber\end{align}
where $D_k = \{d \in D|\mathrm{LangLabel}(d) = k\}$ and $k$~=~$\{$torso, upper arm, forearm, palm, little finger, ring finger, middle finger, index finger, thumb$\}$.

\begin{figure}[!ht]
\centering
\includegraphics[width=450 px,clip]{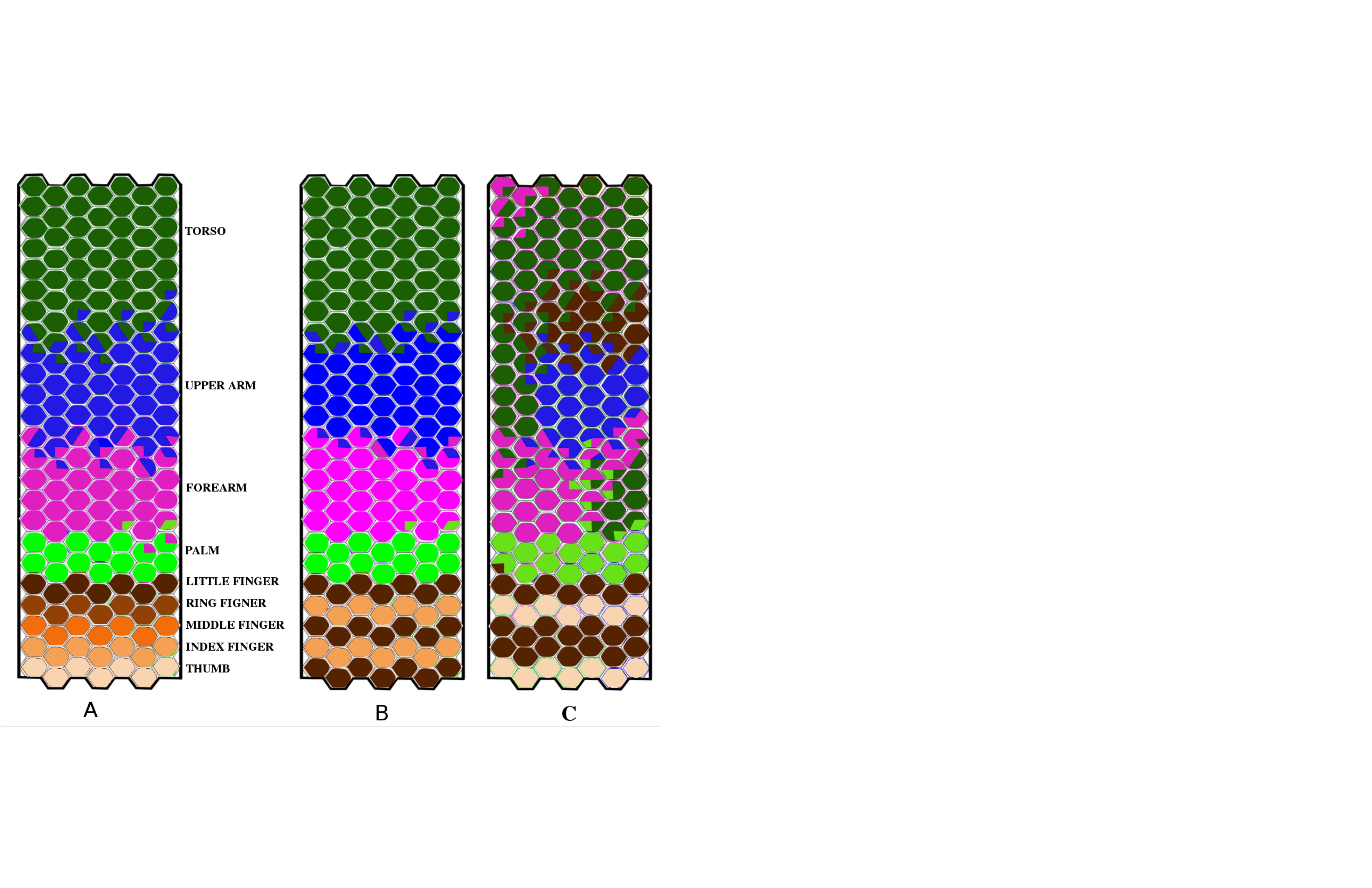}
\caption{Projection of mapping results back onto the tactile homunculus -- sample runs of the algorithm. Color code for individual body parts is the same as in Fig.~\ref{fig:icubHomun}. (A) Original homunculus with true labels. (B) Results from data set with 6381 data points and 10\% noise. (C) Results from 638 data points and 80\% noise. }
\label{fig:homlang}       
\end{figure}

Afterwards, we visualize for each neuron how much it is activated for individual body parts. Results for data sets of differing size and level of noise in the linguistic domain can be seen in Fig.~\ref{fig:homlang}. Clearly, for large enough data sets and limited noise, the mapping from language to the tactile modality is successful in delineating the body part categories (the fingers with fewer data points being more challenging)---as can be seen by comparing panels A and B. 

\section{Discussion and Conclusion}
To study the problem of associating (mapping) between sensorimotor or multimodal information, concepts or categories, and language or symbols, we have chosen a specific but less studied instance of this problem: segmentation and labeling of body parts. Perhaps, from a developmental perspective, this could be plausible, as the body may be the first ``object'' the infant is discovering. The self-exploration occurs in the sensorimotor domain, but at the same time or slightly later, the infant is exposed to utterances of body part names. In this work, we study the mapping between the tactile modality and body part labels from linguistic input. 

We present a new algorithm for mapping language to sensory modalities (sequential mapping), compare it to one-step mapping and test it on the body part categorization scenario.
Our results suggest that this mapping procedure is robust, resistant against noise, and sequential mapping shows better performance than one-step mapping for all data set sizes and also slower performance degradation with increasing noise in the linguistic input. 
Furthermore, we explored accuracy of the sequential mapping for individual body parts, revealing that body parts less represented in the data set---fingers---were categorized less accurately. This problem might be mitigated with increased overall data set size; yet, dealing with clusters with uneven data point number is a  common problem of clustering  algorithms (in our case GMM). 

Projecting the labels or categories induced by language back onto the tactile homunculus showed that the body part categories are quite accurate. Given the nature of the tactile input---the skin is a continuous receptor surface---and the random-uniform tactile input generator used, the linguistic input was the only one that can facilitate cluster formation. However, more realistic, non-uniform touch and, in particular, the addition of additional modalities (proprioception, vision) should enable bottom-up non-linguistic body part category formation, as described by \citep{Vignemont2009}, for example. These constitute possible directions of our future work: the ``modal'' cluster formation will interact with the labels imposed by language. Furthermore, thus far, only one half of the body was considered---corresponding to the lateralized representations in the tactile homunculus---, but one can imagine stimulating both left and right arm, for example, while hearing always the same utterance: `upper arm'. Further study of the brain areas involved in this processing is needed, in order to develop models more closely inspired by the functional cortical networks, like in \citep{Caligiore2010} that model the experimental findings of \citep{Borghi2004}.

For our experiments we used artificially generated linguistic input (i.e., body part labels) with added noise (i.e. wrong labels with a certain probability). In the future, we are planning to use actual auditory input (spoken words) with real noise. This will also add the additional dimension of similarity in the auditory domain: `arm' and `forearm' are phonetically closer to each other than to, say, `torso'. Thus, the linguistic modality will not constitute crisp, discrete labels anymore, but these will have to be extracted first---opening up further possibilities for bidirectional interaction with other modalities.

\section{Acknowledgement}
K.S. and M.H. were supported by the Czech Science Foundation under Project GA17-15697Y. M.H. was additionally supported by a Marie Curie Intra European Fellowship (iCub Body Schema 625727) within the 7th European Community Framework Programme. Z.S. was supported by The Grant Agency of the CTU Prague project SGS16/161/OHK3/2T/13. M.V. was supported by European research project TRADR funded by the EU FP7 Programme, ICT: Cognitive systems, interaction, robotics (Project Nr. 609763).
 
 \bibliography{references}
 \bibliographystyle{apalike_kuz}
%
%
%
\end{document}